\documentclass[letterpaper, 10 pt, conference]{ieeeconf}  

\IEEEoverridecommandlockouts                              
\usepackage{graphicx}
\usepackage{subfigure}
\usepackage{amsmath}
\usepackage{algorithm}
\usepackage{algorithmic}
\usepackage{array}
\usepackage{microtype}
\title{\LARGE \bf
Computationally Efficient Obstacle Avoidance Trajectory Planner for UAVs Based on Heuristic Angular Search Method
}

\author{Han Chen$^1$ and Peng Lu$^{1,2*}$ 
\thanks{$^1$The Adaptive Robotic Controls Lab (ArcLab), Hong Kong Polytechnic University, Hung Hom, Kowloon, Hong Kong, China.
{\tt\small stark.chen@connect.polyu.hk}}
\thanks{$^2$Department of Mechanical Engineering, The University of Hong Kong, Pokfulam, Hong Kong, China.
{\tt\small lupeng@hku.hk}}
\thanks{$^*$Corresponding author}}%

\begin{document}

\maketitle
\thispagestyle{empty}
\pagestyle{empty}

\begin{abstract}

For accomplishing a variety of missions in challenging environments, the capability of navigating with full autonomy while avoiding unexpected obstacles is the most crucial requirement for UAVs in real applications. In this paper, we proposed such a computationally efficient obstacle avoidance trajectory planner that can be used in unknown cluttered environments. Because of the narrow view field of single depth camera on a UAV, the information of obstacles around is quite limited thus the shortest entire path is difficult to achieve. Therefore we focus on the time cost of the trajectory planner and safety rather than other factors. This planner is mainly composed of a point cloud processor, a waypoint publisher with Heuristic Angular Search(HAS) method and a motion planner with minimum acceleration optimization. Furthermore, we propose several techniques to enhance safety by making the possibility of finding a feasible trajectory as large as possible. The proposed approach is implemented to run onboard in real-time and is tested extensively in simulation and the average control output calculating time of iteration steps is less than 18 ms.

\end{abstract}

\section{INTRODUCTION}

	Unmanned aerial vehicles(UAVs), especially quadrotors, are increasingly used in field applications due to their flexibility, agility, and stability. Autonomous navigation enables the aircraft to be used for missions inaccessible or dangerous to humans or ground vehicles, such as search and rescue, inspection and exploration, monitoring and surveillance. For the UAV obstacle avoidance, the most important thing is responding quickly enough to the newly detected static obstacles, or even moving obstacles. First, the trajectory planner needs to obtain information about obstacles in the environment. In most related studies, obstacles are obtained by using two sensors: lidar or depth binocular camera. Lidars are generally large in size and weight and consume too much energy. Although lidars have higher detection accuracy and more stable obstacle information, they are not suitable for small drones. The detection accuracy of the depth camera is sufficient for UAV obstacle avoidance within a certain distance (0.5-8m), but the field of view is narrow, and it is impossible to obtain $360^\circ$ environmental information like lidar. So we need to use the information obtained by the depth camera to build a global map, which can prevent the drone from hitting obstacles outside the field of view. The accuracy and stability of maps built online is key to avoiding obstacles. 

   \begin{figure}[thpb]
      \centering
      \includegraphics[width=0.43\textwidth , height=2.8cm]{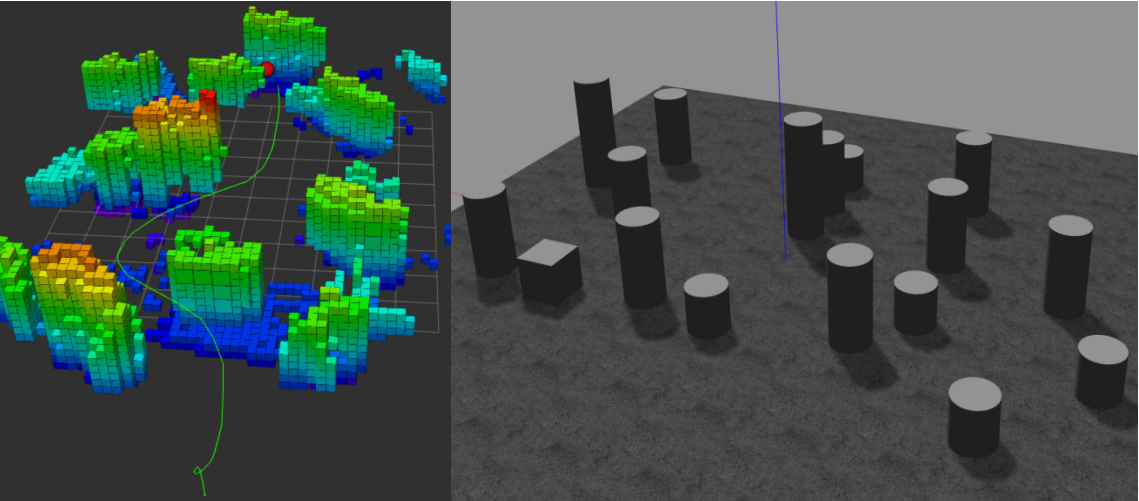}
      \caption{Our simulation environment and visualized data of results}
      \label{figurelabel}
      \vspace{-0.1cm}
   \end{figure}

In this paper, we propose a method to directly find the target point of the drone in the next step on a sparse point cloud, and then solve the optimization problem to obtain the motion primitives that the drone needs to perform at the next moment. In order to reduce the amount of calculation for collision detection when searching for a waypoint, we further streamline the point cloud of obstacles in the global map maintained by Octomap. The degree of simplification is related to the drone safety radius $r_{\!safe}$ we set. Then, the discrete angular search is used to simplify the collision detection to calculate the distance from the point to the straight line. 

In summary, the main contributions of the paper are:

\begin{itemize}

\item The combination of a streamlined point cloud of global Octomap and the heuristic discrete angular search makes the computation load of finding a collision-free path much lighter. It improves the efficiency by generating waypoints directly on the point cloud rather than building a grid map and running a static path planning algorithm(such as A* or JPS) on the grid map afterward.

\item The collision check can be removed from the motion planning part due to the introduction of $r_{\!safe}$ and the constraint of maximum speed and acceleration of drone, the drone's position can be well constrained in the free space between the execution time of two contiguous steps of the trajectory planner.

\item We propose three techniques to guarantee safety in the autonomous flight based on the mentioned method. Simulation experiments in ROS/Gazebo showing agile flights in completely unknown cluttered environments, with maximal average control output calculating time of iteration steps less than 18 ms.
\end{itemize}

\section{RELATED WORK}


For hardware experiments, it is necessary to encode and use the information of detected obstacles in an efficient way. In most of the related research, point cloud is the most widely used form to express obstacle information. For the use of point clouds, the most common practice is to use a filtered point cloud to create a three-dimensional grid map and then perform trajectory planning on the basis of the grid map[1]. Considering the estimation of the vehicle state, many methods have been proposed to convert the depth measurements generated by the on-board sensors into a global map. Representative methods include voxel grids [2], Octomap [3], and elevation maps [4]. Each method has advantages and disadvantages in a particular environment. The voxel grid is suitable for fine-grained representation of small volumes, but the storage complexity is poor. Elevations are suitable for representing artificial structures composed of vertical walls but are less efficient in describing natural and unstructured scenes. Octomap is memory-efficient when indicating an environment with a large open space. This storage structure is very useful for further utilizing maps for trajectory planning and has the function of automatic map maintenance, which is convenient to use and has satisfactory results in both simulation and hardware flight tests.

In the previous work [5], they used Octomap building on point cloud raw data to develop their own method and gained good experimental results. In another way, in order to reduce the computational time consumed by this step of building the map, some researchers have directly planned the trajectory on the original point cloud. Lopez used the transformed point cloud for the collision check with trajectories corresponding to the randomly generated motion primitives [6]. However, planning on the point cloud directly requires high-quality point cloud information, and this method is not suitable for drones carrying a single depth camera if a global map has not been established.

After obtaining the environmental information, the most important thing is to calculate motion primitives. The related methods can also be divided into two categories. One is to first convert the obstacle information and the position of the UAV in three-dimensional space into a local map. This map contains only the obstacle information near the UAV, the global goal and the points of obstacles are projected in this local map in some way, then a static path planning algorithm is run on the local map, and finally the motion primitives are obtained by solving the motion planning equations. For instance, [7]-[8] built a local occupancy grid map with the most recent perception data and generated a minimum-jerk trajectory through waypoints from an A* search. As for waypoint time allocation, an approximate method was used in [9] and a bi-level optimization was used  in [10]-[11] to find the times. The other type of method is to skip searching paths on the map first and directly generate motion primitives by sampling. Then, the evaluation function can be designed to select the most suitable group of motion primitives as the output, which is very similar to DWA. A representative work is presented by Mueller et al, even making the quadrotor catch a falling ball [12].

In addition, you can also directly obtain motion primitives by solving an optimization problem. This requires appropriate expressions of the trajectory of the aircraft, such as Bezier curves, and to ensure that the final trajectory is collision-free by setting constraints. [13]-[15] achieved satisfactory results by utilizing this method. For these two methods, the collision check is the most time-consuming, and it is difficult to significantly increase the calculation speed within its own framework, so we propose another idea to improve the calculation speed.

\section{QUICK RESPONDING AND SAFE PLANNER}
\vspace{-0.3cm}
\begin{figure}[thpb]
\centering
\subfigure[]{
\includegraphics[width=0.15\textwidth ,height=3cm]{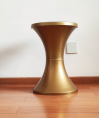}}
\hfill
\centering
\subfigure[]{
\includegraphics[width=0.15\textwidth ,height=3cm]{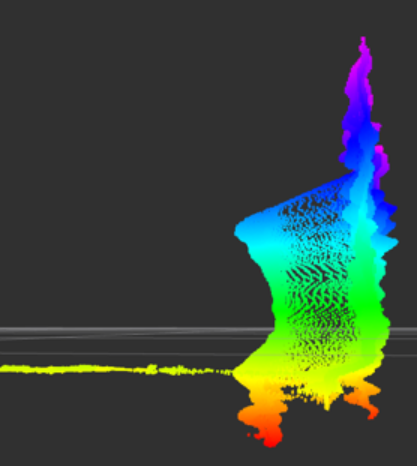}}
\hfill
\centering
\subfigure[]{
\includegraphics[width=0.15\textwidth ,height=3cm]{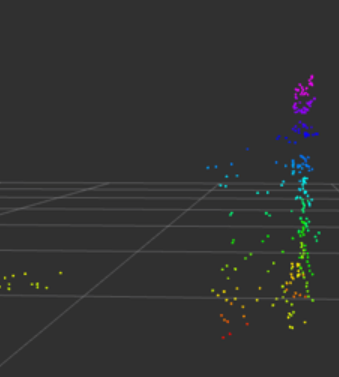}}
\caption{(a) depth camera’s RGB output, (b) raw point cloud, (c) filtered point cloud($Pcl_{2}$)}
\vspace{-0.2cm}
\end{figure}
As mentioned above, the collision check is the most time-consuming part of the trajectory generation. To cope with this challenge, we introduce a Heuristic Angular Search(HAS) method with a backup safety plan. The overall algorithm is presented in {\bf Algorithm 1}, where $Pcl$ is the point cloud, $P_{rec}$ is the list where the planner record $p_n$ in each step, $ B_E$ described in (1) is the transformation matrix from body coordinate to earth coordinate, $c()$ is short for $cos()$ and $s()$ is short for $sin()$, $\phi, \theta ,\psi$ are Euler angles respectively. We describe Line 2-4 in section A and describe Line 5 in Section B. Line 6-7 is described in Section C and Line 9 is described in Section D. 
Overall, the outer loop can be executed at 55-100 Hz, considering the density of obstacles in the simulation tests.

\begin{algorithm}[h]
\caption{our proposed planner} 
\label{alg1}
\begin{algorithmic}[1]
\WHILE{$true:$} 
\STATE Filter the raw point cloud data, output \textbf{$Pcl_1$} 
\STATE Transform $Pcl_1$ in body coordinate (B) to $Pcl_2$ in earth coordinate(E) by $B_E$ 
\STATE Build a global map represented by point cloud \textbf{$Pcl_3$}, filter again 
\STATE Find the next waypoint $w_p$ by heuristic angular search
\IF{$\textbf{found a feasible waypoint:}$} 
\STATE Run the minimum acceleration motion planner to get motion primitives
\ELSE 
\STATE Run the backup plan for safety, then go to 5
\ENDIF
\STATE Send the motion primitives to the UAV flight controller
\STATE Record the current position $p_n$ in list $P_{rec}$
\ENDWHILE
\end{algorithmic}
\end{algorithm} 
\vspace{-0.2cm}

$$ B_{E}=\left[\begin{array}{ccc}
\mathrm{c} \psi \mathrm{c} \theta & \mathrm{s} \psi \mathrm{c} \theta & - \mathrm{s} \theta\\
\mathrm{c} \psi \mathrm{s} \theta \mathrm{s} \phi-\mathrm{s} \psi \mathrm{c} \phi & \mathrm{s} \psi \mathrm{s} \theta \mathrm{s} \phi+\mathrm{c} \psi \mathrm{c} \phi & \mathrm{c} \theta \mathrm{s} \phi \\
\mathrm{c} \psi \mathrm{s} \theta \mathrm{c} \phi+\mathrm{s} \psi \mathrm{s} \phi & \mathrm{s} \psi \mathrm{s} \theta \mathrm{c} \phi-\mathrm{c} \psi \mathrm{s} \phi & \mathrm{c} \theta \mathrm{c} \phi
\end{array}\right] \eqno{(1)} $$
\vspace{-0.2cm}

\subsection{Processing the point cloud}
The point cloud data obtained by a real depth camera is often noisy and too dense, and the noise is greater on objects farther from the camera, as shown in Fig. 2(a) and Fig. 2(b). This is inconvenient for converting the coordinate system of each point in the point cloud and establishing a global map. First, we filter the original point cloud data $Pcl_1$ through three filters in order to obtain the point cloud data $Pcl_2$ which is convenient to store and recall. The algorithm of the filter and the point cloud after filtering are shown in {\bf Algorithm 2} and Fig. 2(c). $ d_{use}$ is a parameter. It can be seen that the filtered point cloud data are more concise and tidy, retaining the basic shape of the obstacle. Then we convert the point cloud into the earth coordinate system and use Octomap to build and maintain a global map. In fact, it is tolerable as long as the gap between the midpoints of the point cloud corresponding to an obstacle is not greater than the safe radius  $ r_{\!safe}$  of the drone. But if you do this at beginning, the global map after fusion will be unavailable for visualization.  So we filter again after we obtain the point cloud of the global map, the algorithm is also shown in \textbf{Algorithm 2}. $q$ is one of the three axes' value of a point in $Pcl_4$, $point_w$ is the point in $Pcl_4$, $L_q$ is the list of all points in $Pcl_4$ and it is rearranged by $q$ value according to the order of $x-y-z$.

At last, we only use the point in $Pcl_5$ for collision detection.

\begin{algorithm}[h]
\caption{point cloud filter} 
\label{alg2}
\begin{algorithmic}[1]
\STATE $Pcl_{1}  \Leftarrow $ point cloud raw data
\FOR{$point_i$ in $Pcl_{1}$}	
\STATE Remove $point_i$ which is further than 8m, keep only one point in a 0.2m voxel, remove the outliers
\ENDFOR 
\STATE $Pcl_{2} \Leftarrow Pcl_1$
\FOR{$point_{m}$ in $Pcl_{2}$}
\STATE $point_{m}= point_{m}+p_n$
\ENDFOR
\STATE $Pcl_{3} \Leftarrow$ center points of Octomap, with $Pcl_2$ input
\STATE $Pcl_{4} \Leftarrow Pcl_{3}$
\FOR{$q$ in ${x,y,z}$}
\FOR{$q_{w}$ of $point_{w}$ in $L_q$}
\IF{not $((q_{w}-L_{q}(0,q))\% r_{\!safe}\approx 0$
or no element of $L_q(:,q)$ in range of $[q_{w} ,q_{w} + r_{\!safe}])$} 
\STATE Delete $point_{w}$ from $Pcl_4$
\ENDIF
\ENDFOR
\ENDFOR
\STATE $Pcl_{5} \Leftarrow Pcl_{4}$
\FOR{$ point_{t} $ in $Pcl_5$}	
\IF{$\|\overrightarrow{p_{n}point_{t}}\|_{2} >d_{use}$}
\STATE Delete $point_t$ from $Pcl_{5}$
\ENDIF
\STATE $d_{min}=min(\|\overrightarrow{p_{n}point_{t}}\|_{2})$
\ENDFOR
\end{algorithmic}
\end{algorithm} 
\vspace{-0.0cm}

\subsection{Heuristic angular search method}

Different from the previous work in which a complete path needs to be planned on the local map, we only find a target point close to the drone as a guide for motion planning. Because the overall planner's calculation speed is quite fast, such a short predicted trajectory is sufficient to refresh before the drone flight reaches its endpoint. As shown in Fig. 3, we use the vector $A_{g} = (\alpha_g, \beta_g)$ to represent the angle of the navigation target $G = (x_g, y_g, z_g)$ relative to the current position of the drone $p_{n}(x, y, z)$ in $E$. Based on this, we define a series of line segments with different endpoints $P_{d1}$-$P_{d4}$ in (2), and these line segments have a common endpoint $p_n$.

\textbf{Algorithm 3} reveals the process of searching for waypoints. We simplify the calculation of collision detection by calculating the perpendicular distance from a point to a line segment, rather than the distance from the obstacle to the sampled curvy path [16]. The specific process of collision checking is shown in \textbf{Algorithm 4}. In most cases in simulation tests, collision detection can be done within 16 ms. The meaning of heuristic search is that the starting point of the search is calculated according to the historical record of the results obtained by this method and the current point cloud information, so as try to obtain an initial value $A_{g0}$ which is the closest to the final search result, minimize the search time cost. The initial value of $A_{g0}$ is calculated in (3) and (4), where $l_d$ is the detection radius of UAV for obstacle avoidance check, $\mu$ is a relatively small coefficient with a value between 0.1-0.2. $A_{last}$ is the angle corresponding to the waypoint in the last step, $n_{obs}$ is the size of $Pcl_5$ in the current step and $n_{avr}$ is the average of the size of $Pcl_5$ over all past steps.

\vspace{-0.4cm}
$$\begin{aligned}
&P_{d 1}=l_{d}\left(\mathrm{c}\left(\alpha_{g 0}+\alpha_{d}\right), \mathrm{s}\left(\alpha_{g 0}+\alpha_{d}\right), \mathrm{s}\left(\beta_{g 0}\right)\right)+p_n\\
&P_{d 2}=l_{d}\left(\mathrm{c}\left(\alpha_{g 0}-\alpha_{d}\right), \mathrm{s}\left(\alpha_{g 0}-\alpha_{d}\right), \mathrm{s}\left(\beta_{g 0}\right)\right)+p_n \\
&P_{d 3}=l_{d}\left(\mathrm{c}\left(\alpha_{g 0}\right), \mathrm{s}\left(\alpha_{g 0}\right), \mathrm{s}\left(\beta_{g 0}+\alpha_{d}\right)\right)+p_n \\
&P_{d 4}=l_{d}\left(\mathrm{c}\left(\alpha_{g 0}\right), \mathrm{s}\left(\alpha_{g 0}\right), \mathrm{s}\left(\beta_{g 0}-\alpha_{d}\right)\right)+p_n
\end{aligned}\eqno{(2)}$$

$$
A_{g 0}=\left\{\begin{array}{l}
\left.A_{g} \text { (others }\right) \\
A_{\text {last}}\left(\lambda n_{\text {obs}}>n_{avr}\right)
\end{array}\right. \\
\eqno{(3)}$$
$$\lambda=\frac{\text{times' number for } A_{g}=A_{last} \ \text{in last } 3 \text { steps }}{3} \eqno{(4)}$$

\begin{figure}[thpb]
\centering
\subfigure[]{
\includegraphics[width=0.45\textwidth , height=4.5cm]{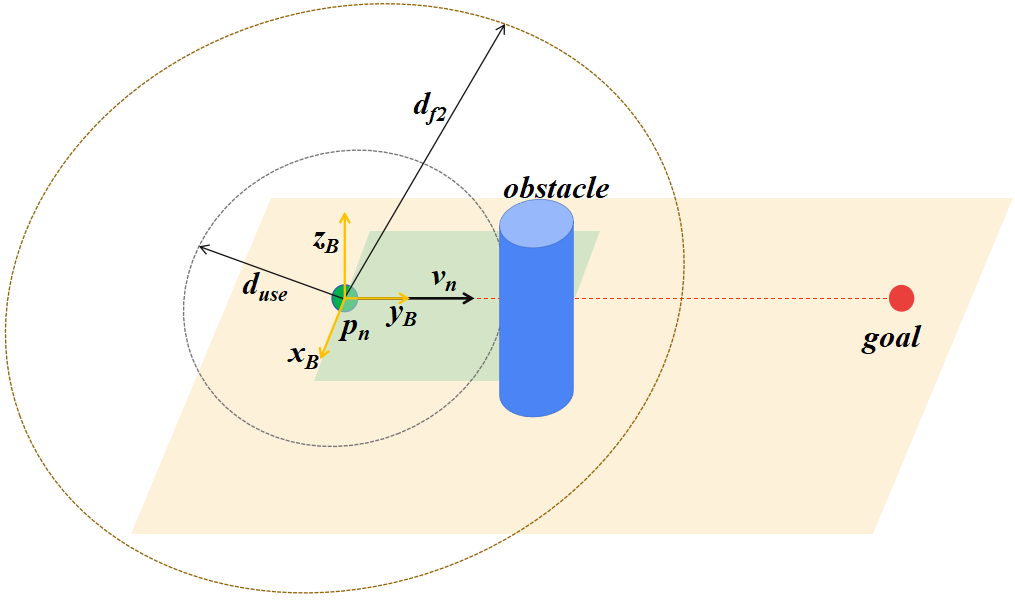}}
\hfill
\centering
\subfigure[]{
\includegraphics[width=0.23\textwidth]{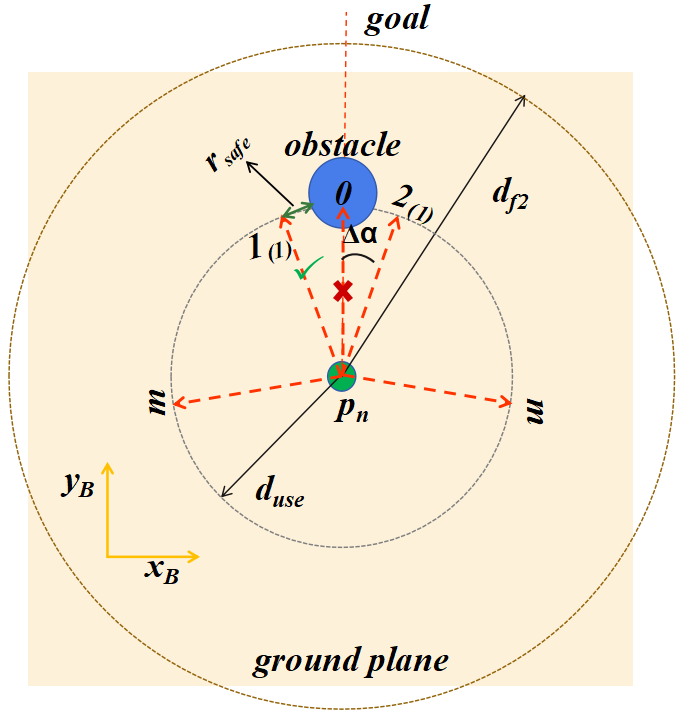}}
\hfill
\centering
\subfigure[]{
\includegraphics[width=0.23\textwidth]{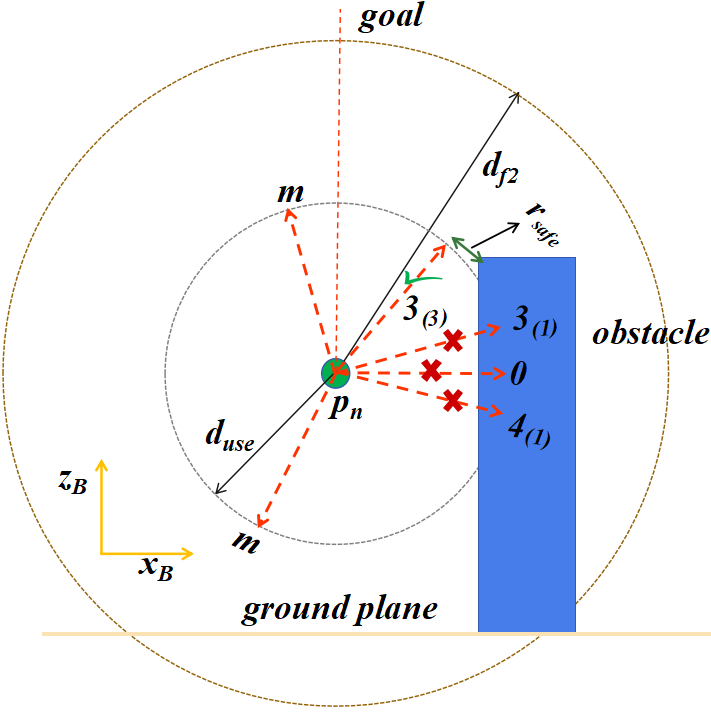}}
\caption{Illustration about angular search,(a) is a stereogram,(b) and (c) are the projection of (a)  to different plane. ${B-xyz}$ presents the body coordinate. The number in the blanket is the ordinal number of iteration, for example, $2_{(1)}$  presents $P_{d2}$ with $\alpha_d=\Delta\alpha$.}
\vspace{-0.5cm}
\end{figure}

\begin{algorithm}[thpb]
\caption{HAS method} 
\label{alg3}
\begin{algorithmic}[1]
\FOR{$i$ in {1,2,3,4}}
\FOR{$\alpha_d$ in ${0,\Delta\alpha,2\Delta\alpha...m\Delta\alpha}$}
\IF{$d_{Pdi} >r_{\!safe}$} 
\STATE $w_{p}=\mu(P_{di}-p_{n}) + p_n$
\STATE Break all circle
\ENDIF
\ENDFOR
\ENDFOR
\end{algorithmic}
\end{algorithm} 
\vspace{-0.2cm}

\begin{algorithm}[tbh]
\caption{collision check} 
\label{alg4}
\begin{algorithmic}[1]
\FOR{$point_t$ in $Pcl_5$}
\IF{$\|\overrightarrow{p_{n}point_{t}}\|_{2}^{2} \textgreater
\|\overrightarrow{P_{di}point_{t}}\|_{2}^{2}$+$\|\overrightarrow{P_{di}p_{n}}\|_{2}^{2}$\\
or $\|\overrightarrow{P_{di}point_{t}}\|_{2}^{2} \textgreater \|\overrightarrow{p_{n}point_{t}}\|_{2}^{2}$+$\|\overrightarrow{P_{di}p_{n}}\|_{2}^{2}$} 
\STATE $d_{Pdi}=\infty$ (Foot drop of $point_t$ is not on $\overline{p_{n}P_{di}}$) 
\ELSE
\STATE $d_{Pdi}=\frac{\|\overrightarrow{p_{n} point_t} \times \overrightarrow{p_{n} P_{di}}\|_{2}}{\left\|\overrightarrow{p_{n} P_{di}}\right\|_{2}}$
\ENDIF
\ENDFOR
\end{algorithmic}
\end{algorithm} 

\subsection{Motion planning}
Once the path point has been obtained, the next step is to calculate the control command, such as position $p =(p_x, p_y, p_z )$, speed $v= (v_x, v_y, v_z )$, acceleration $a= (a_x, a_y, a_z )$, and send the command to the flight controller, so as to ensure that the aircraft can fly within its own kinematic limit and reach the next waypoint. Generally, the motion primitives are obtained by solving an optimization problem. In this way, the kinematic constraints of the drone can be addressed by setting constraints [17]. We take the acceleration of the drone as the variable to be solved, because compared with the use of jerk or snap, acceleration can be directly sent to the flight control as a control command, and the computational load is less while meeting the kinematic constraints and ensuring the smooth trajectory. 

The optimization problem is defined in (5), where the subscript $n$ presents the current step in a rolling process of the whole planner, $p_{start}$ is the position of the drone when the planner starts to work [18]. $v_{max}$ and $a_{max}$ are the kinematic constraints for speed and acceleration respectively, $t_{max}$ is the upper bound for the time which can be used to finish the predicted piece of trajectory. $\xi$ is the tolerance for the difference between the end of the predicted trajectory and the $w_p$, $v_{n+1}$ and $p_{n+1}$ are calculated by the kinematic formula.

\vspace{-0.0cm}
$$\begin{aligned}
\min _{a_{n}, t_{n}} \quad &\left\|a_{n}\right\|_{2}^{2}+\eta t_{n}\\
\text {s.t.}\quad & p_{0}=p_{\text {start}} \\
&0< \mathrm{t}_{n} \leq t_{max } \\
&v_{n} =\dot{p}_{n} \\
&a_{n} =\dot{v}_{n}\\
&\left\|v_{n+1}\right\|_{\infty} \leq v_{max}\\
&\left\|a_{n}\right\|_{\infty} \leq a_{max}\\
&\| p_{n+1}- w_{p} \|_{2} \leq \xi\\
&v_{n+1}=v_{n}+a_{n} t_{n}\\
&p_{n+1}=p_{n}+v_{n} t_{n}+\frac{1}{2} a_{n} t_{n}^{2}
\end{aligned} \eqno{(5)}$$

\subsection{Safety guarantee}
Next, we demonstrate the safety of the trajectory and add additional measures to improve safety based on the above method. As shown in Fig. 4, if the trajectory of the aircraft is a straight line that coincides with $\overline{p_{n}w_p}$ in each step, then this line must be safe because it has undergone collision detection check. However, considering the kinematic constraints of the aircraft, the trajectory of the aircraft in each step is a curve. Assuming that the acceleration $a_n$ solved by the optimizer is in the same plane as the speed $v_n$ and the waypoint of the drone at the current moment (so that it meets the optimization objective function), then this curve is a parabola in this plane. When $a_n$ is in opposite direction of $v_n$, the deviation $d_{max}$ between the drone trajectory and line segment $\overline{p_{n}w_p}$ is the largest. It can be easily proven since $\Vert v_{n} \Vert_2 $ and $\Vert a_{n} \Vert_2 $ is constant.

We can get $d_{max}$ by solving the optimization problem in (6), and we get a close form solution in (7).

$$\begin{aligned}
&d_{\max }=\max \left(\frac{2\left\|v_{n}\right\|_{2}^{2}}{\left\|a_{n}\right\|_{2}}\right)\\
s.t.\ &\sqrt{\frac{2\left\|w_{p}-p_{n}\right\|_{2}}{\left\|a_{n}\right\|_{2}}}+\frac{\left\|v_{n}\right\|_{2}}{\left\|a_{n}\right\|_{2}} \leq t_{\max }
\end{aligned} \eqno{(6)}$$

$$d_{\max }=2\left\|v_{n}\right\|_{2}\left(t_{\max }-\sqrt{\left.\frac{2\left\|w_{p}-p_{n}\right\|_{2}}{a_{\max }}\right.}\right)\eqno{(7)}$$

The trajectory is safe when we choose parameters to make $d_{max} < r_{\!safe}$. Besides, we implement another three techniques to achieve further security:

1)change $l_d$ to a smaller one when no feasible $w_p$ is found at the first circle and run another circle.

2)change $v_{max}$ to a smaller one when $d_{min}$ is smaller than 1.5$r_{\!safe}$ [19].

3)if no feasible $w_p$ is found, return to the last path point and use the next feasible solution in the angular search.


   \begin{figure}[thpb]
      \centering
      \includegraphics[width=0.42\textwidth]{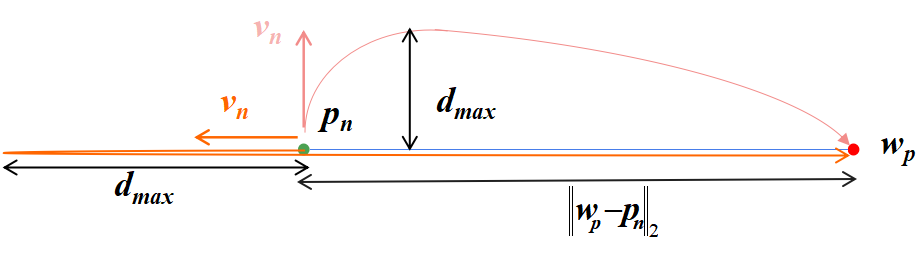}
      \caption{Illustration of the relationship between $d_{max}$ and the direction of $v_n$ ($\| {v_n}\|_2$ and  $\|a_n\|_2$ are fixed).}
      \label{figurelabel}
      \vspace{-0.0cm}
   \end{figure}
   
\section{EXPERIMENTAL RESULTS}
\subsection{Experimental Configuration}

Our proposed HAS-based trajectory planner was tested and verified in the Robot Operation System (ROS)/Gazebo simulation environment. Gazebo is a simulation software which can provide a physical simulation environment  close to the real world.  The model of the drone we use in the simulation is IRIS, the depth camera model is Kinect V2, and the PX4 1.7.4 firmware version is used as the underlying flight controller. Mavros package is deployed for establishing the communication between our planner node and the PX4 control module. The acceleration controller for tracking is provided by the PX4 module by default.  The point cloud processor is executed by C++ code and the other parts are executed by Python scripts. All these timing breakdowns were measured using an IntelCore i7-8200U 1.8GHz Processor. Table \uppercase\expandafter{\romannumeral1} describes the parameter settings of the planner in the simulation test. To make the depth camera observe the environment more efficiently, we control the yaw angle of the drone to keep the camera always heading toward the goal during the flight.
\begin{table}[thpb]
\setlength{\abovecaptionskip}{-8 pt}
\caption{PARAMETERS FOR SIMULATION}
\label{table_1}
\begin{center}
\begin{tabular}{|c|c|c|c|}
\hline
Parameter& Value &Parameter & Value\\
\hline
$ \Delta\alpha$ & $10^\circ$ &$d_{use}$ &3m\\
\hline
$a_{max}$ & $4m/s^2$ &$r_{\!safe}$ &0.8m\\
\hline
voxel size & $0.2m$ &$v_{max}$ &$3m/s$\\
\hline
$\xi$ & 0.01m &$\mu$ &0.1\\
\hline
$ l_d$ & $3m$ &$d_{max}$ &$0.68m<r_{\!safe}$\\
\hline
$ t_{max}$ & $0.5s$ &$\eta$ &1.2\\
\hline
\end{tabular}
\end{center}
\vspace{-0.5cm}
\end{table}

Two flight tests of increasing difficulty are presented in section B and section C to show the planning trajectory in 3D space and the time cost of each planning step. The obstacles are not known a priori and are unobservable at takeoff.
\subsection{Simulation flight test in a simple environment} The test results are shown in the Fig. 6. In the first flight, the starting point of the drone is (0,0,0), and the red point indicates the navigation target point (12,0,1). After reaching the target point, set the starting point change to (12,0,0) and the endpoint is set to (0,0,1), then another test is performed. This is to test the drone's ability to avoid obstacles in the horizontal and vertical directions. According to the design of the algorithm, the drone will choose the path with the smallest amount of angle change of the flight direction when the obstacle can be avoided both horizontally and vertically. Because turning the drone too fast will increase the noise of the point cloud data obtained by the depth camera and destroy the established map, the attitude angle of the drone should be kept as smooth as possible.

The global 3D map and the flight trajectory of the drone during the flight are displayed in RVIZ, as shown in Fig. 6(c)-(d). In the first flight, the drone first raised its height to avoid the obstacles in the face of short obstacles and then chose to fly to the left to avoid the higher obstacles. In the second test, the drone first chose to fly to the left and then chose to continue to the left, because this minimizes the amount of angle change in the flight direction.
\begin{figure}[thpb]
\centering
\subfigure[]{
\includegraphics[width=0.232\textwidth , height=2.1cm]{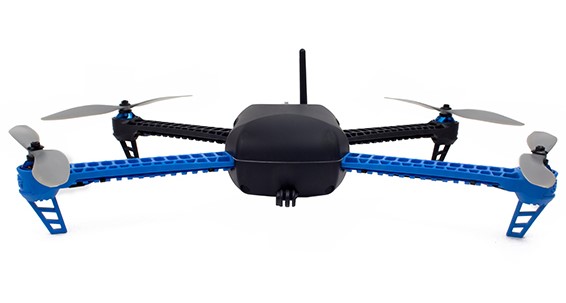}}
\hfill
\centering
\subfigure[]{
\includegraphics[width=0.232\textwidth , height=2.1cm]{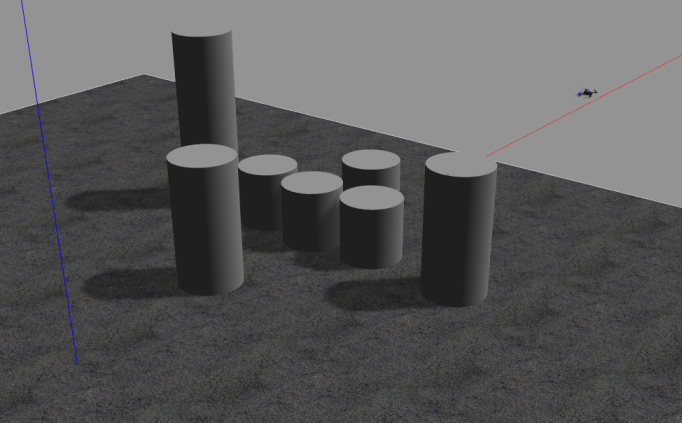}}
\hfill
\centering
\subfigure[]{
\includegraphics[width=0.222\textwidth , height=2.1cm]{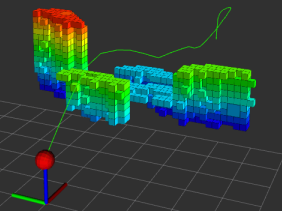}}
\hfill
\centering
\subfigure[]{
\includegraphics[width=0.242\textwidth , height=2.1cm]{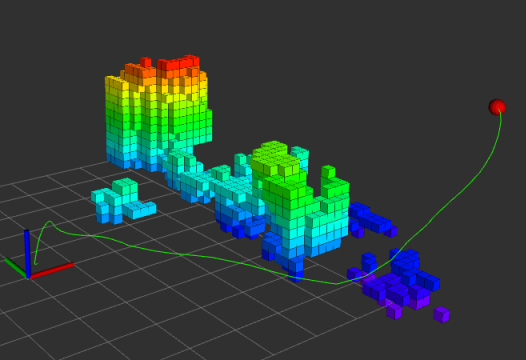}}
\caption{(a) an IRIS drone, (b) a simple simulation environment, (c) and (d) show the results of the first and the second flight test respectively. The trajectory is shown in the green line.}
\vspace{-0.5cm}
\end{figure}

\subsection{Simulation flight test in a complex environment}
In this test, we built a more complex map. Due to the limited space in the paper, we show only one flight's results. The start point is set at (12,0,0), the endpoint is set at (-12,0,1). The flight trajectory of the drone and the established global map are shown in Fig. 7. After repeating the flight experiments 10 times, the detailed data of the trajectory and the average running time of each part of the planner are shown in Fig. 7. Table \uppercase\expandafter{\romannumeral2} compares the calculation time of our proposed planner to the  state-of-the-art, where  Cond.1 means $A_{g0}$ is fixed to goal direction, and Cond.2 means all points in $Pcl_3$ are used for collision check. PL(path length) factor is the ratio of the whole path length to the straight distance from the start point to goal. It is worth noting that the difficulty of the simulation tests in this table are different, we show the data here for a preliminary comparison. We can see that our proposed planner has obvious advantages in computational time, but the whole path length is longer than some of others' works.

\begin{figure}[thpb]
\centering
\subfigure[]{
\includegraphics[width=0.4\textwidth]{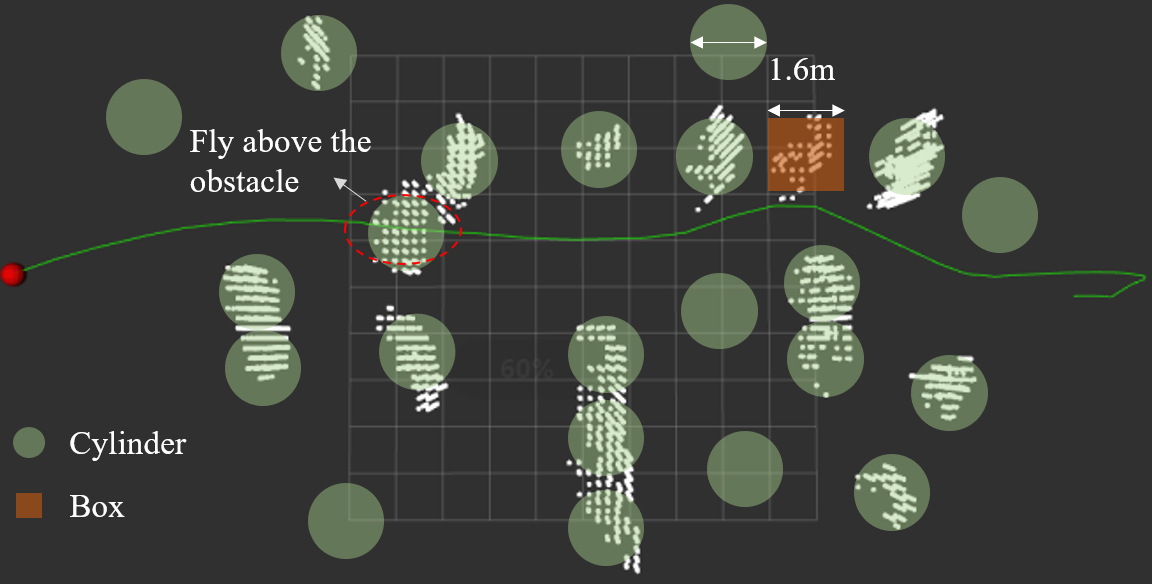}}
\hfill
\centering
\subfigure[]{
\includegraphics[width=0.4\textwidth]{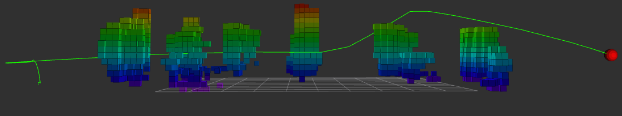}}
\caption{ Results of the test in a complex environment. (a) shows the point cloud of the global map and the size of the obstacles from the top view, (b) shows the Octomap from a side view. The trajectory is shown in the green line.}
\vspace{-0.2cm}
\end{figure}

\begin{figure}[thpb]
\centering
\subfigure[]{
\includegraphics[width=0.4\textwidth , height=2.3cm]{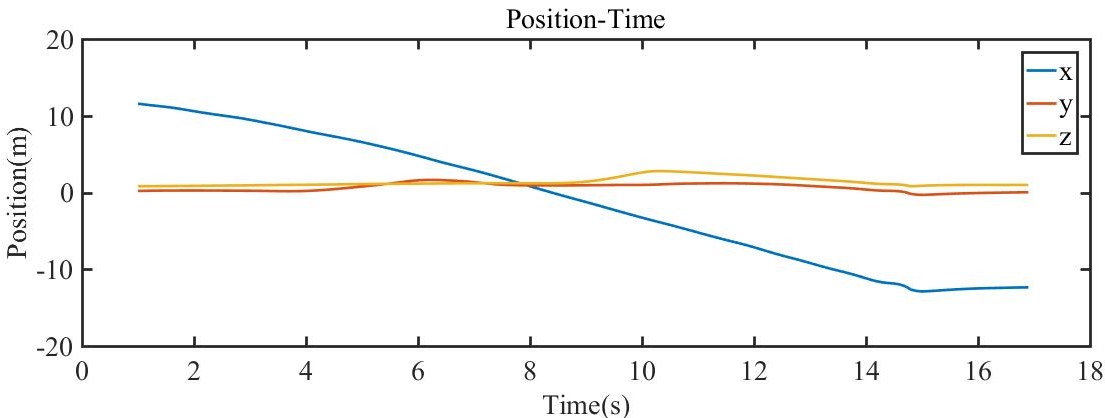}}
\hfill
\centering
\subfigure[]{
\includegraphics[width=0.4\textwidth , height=2.3cm]{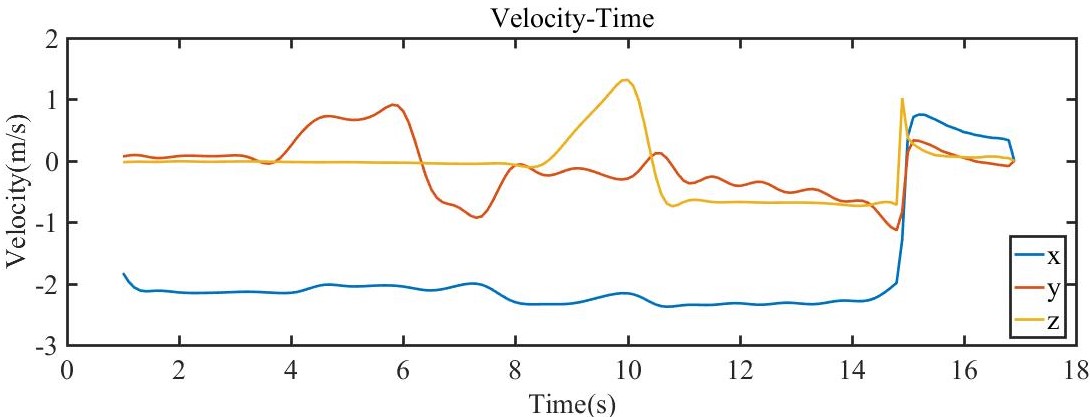}}
\hfill
\centering
\subfigure[]{
\includegraphics[width=0.4\textwidth , height=2.3cm]{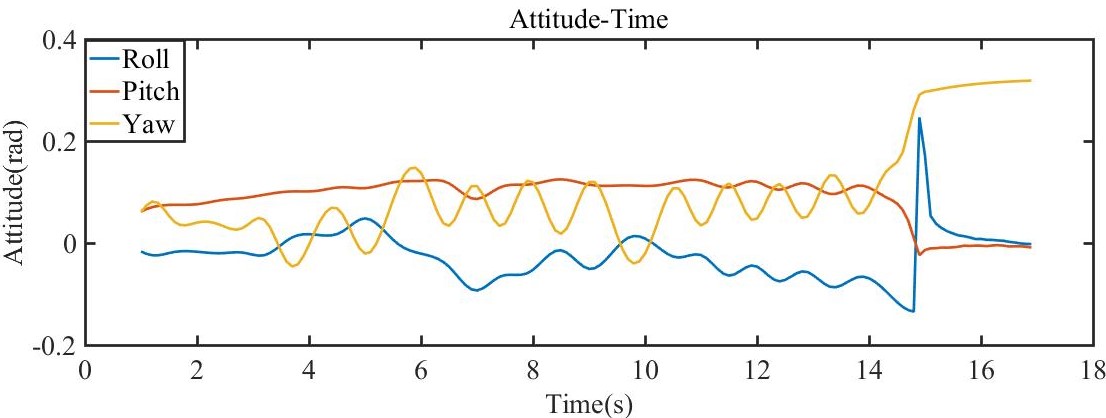}}
\hfill
\centering
\subfigure[]{
\includegraphics[width=0.4\textwidth , height=2.3cm]{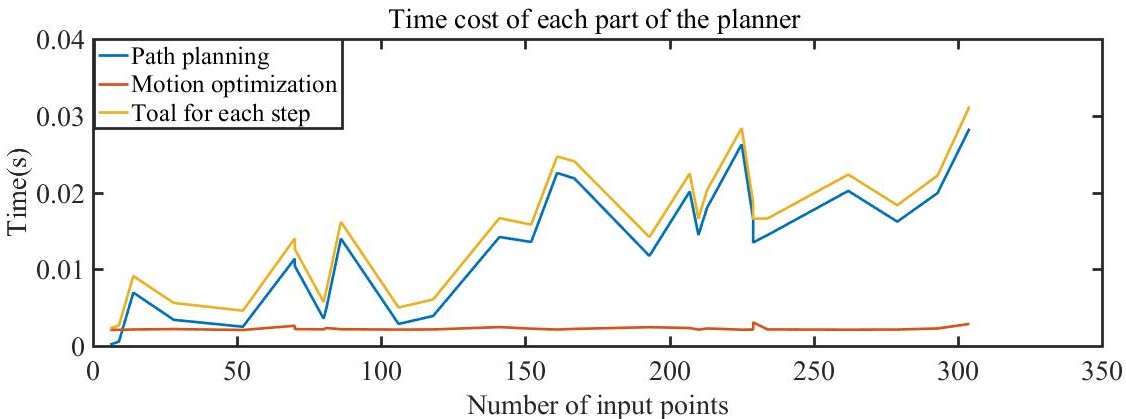}}
\hfill
\centering
\subfigure[]{
\includegraphics[width=0.21\textwidth]{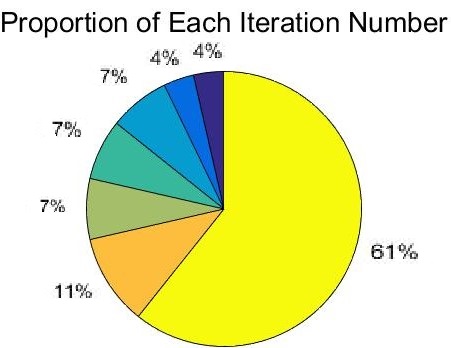}}
\hfill
\centering
\subfigure[]{
\includegraphics[width=0.21\textwidth]{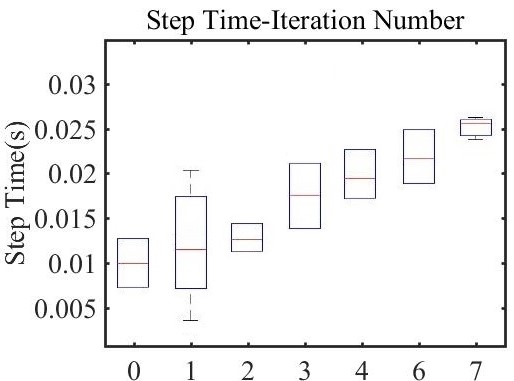}}
\caption{(a)-(c) curve of the three-axis coordinate position, flight speed,  attitude angle respectively; (d) curve of time cost of each part of the planner versus number of points in $Pcl_5$; (e) pie chart for the proportion of each iteration number; (f) the boxplot of time cost for each iteration number.}
\vspace{-0.5cm}
\end{figure}

\begin{figure}[thpb]
\centering
\subfigure[]{
\includegraphics[width=0.2\textwidth]{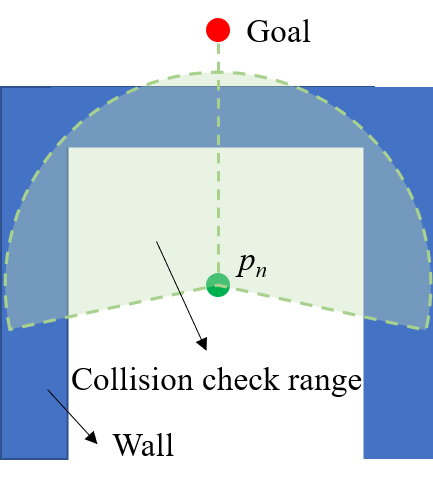}}
\hfill
\centering
\subfigure[]{
\includegraphics[width=0.2\textwidth]{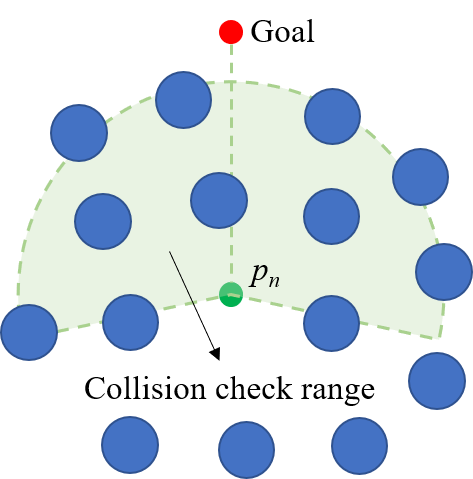}}
\caption{ The two scenarios in which the drone are most likely fail to find a free path}
\vspace{-0.3cm}
\end{figure}

\begin{table}[thpb]
\setlength{\abovecaptionskip}{-8 pt}
\setlength{\belowcaptionskip}{0.5cm}
\caption{COMPARISON WITH STATE-OF-THE-ART ALGORITHMS.}
\label{table_2}
\begin{center}
\begin{tabular}{|p{1.05cm}<{\centering}|p{1.03cm}<{\centering}|p{0.84cm}<{\centering} |p{1.18cm}<{\centering}|p{1.03cm}<{\centering}|p{0.84cm}<{\centering}|}
\hline
Authors & $Comp.$ Time(ms) & PL Factor & Authors & $Comp.$ Time(ms) &PL Factor\\
\hline
Zhou et al.[14] & $>$100 & 1.56 &Tordesillas et al.[22] &$>$25 & \textbf{1.34}\\
\hline
Liu et al.[7] & $>$160 & - &Chen et al.[21] &$>$34 & -\\
\hline
Burri et al.[20] & $>$40 & 1.78&\textbf{This paper} &\textbf{19} & 1.48\\
\hline
This 
paper
(Cond.1) & 29 & 1.48 & This 
paper
(Cond.2) &42 & 1.44\\
\hline
\end{tabular}
\end{center}
\vspace{-0.2cm}
\end{table}
In Fig. 7(b)-(c), the curve changes intensely near the end because the drone was switched to position control mode when it is close enough(<0.3 m in this test) to the goal. We can see from the boxplot that the number of iteration time in the angular search is the major influential factor to the time cost. Fig. 7(e) shows that in most instances the HAS method can work out the feasible solution with less than 3 steps, so the average step time can be controlled within 20ms. The time cost also relates to the number of input points to some extent, which means we can decrease the time cost by simplifying the point cloud($Pcl_5$) in a more efficient way. 

However, in additional series of simulation tests, we found the proposed trajectory planner may fail in two typical scenarios as shown in Fig. 8: a room with narrow exit and a forest with dense and tall pillars. To improve the computation efficient, the collision check is not performed in range of $360^\circ$ on the point cloud. Although the drone returns to the last recorded position when it fails to find a waypoint, it may fail to exit the room if the exit is too narrow. In the dense forest, although the gap between the pillars allows the drone to pass, considering the fixed $r_{\!safe}$ the drone can't find a collision-free line segment under such condition. 

\section{CONCLUSION AND FUTURE WORK}

This work presented a trajectory planner’s framework based on the HAS method, for safe and quick responding flights in unknown environments. The key properties of this planner are that it uses a direct waypoint search method on a simplified point cloud to reduce the time cost and the safety is ensured by restricting $d_{max} < r_{\!safe}$ by setting parameters and compromise on $v_{max}$ and $l_d$ when necessary. Our proposed planner was tested successfully in different simulation environments, achieving the average step time cost within 18 ms. The time cost is believed to be able to achieve a better level on a higher performance hardware platform with C++ code.

In the future, one interesting research idea is to combine the global planner on the global map with the local planner mentioned in this paper together to confront a more complex and dense environment, which is difficult for our current local planner with HAS method to pass. Another potential plan is studying on how to make a drone fly a optimal trajectory in an environment with dynamic and static obstacles at the same time.

\addtolength{\textheight}{-2cm}   





\end{document}